\UseRawInputEncoding
\documentclass[conference]{ieeeconf}
\usepackage{amsmath}
\usepackage{xcolor, soul}
\sethlcolor{yellow}
\usepackage{threeparttable}
\usepackage{multirow}
\usepackage{graphicx}
\usepackage{algorithm}
\usepackage{algpseudocode}
\usepackage{subfigure}
\usepackage{tabularx}
\usepackage{xcolor}
\usepackage{booktabs}
\usepackage{balance}
\usepackage{siunitx}
\usepackage{layouts}
\usepackage{cite}
\usepackage{amsmath}
\usepackage{amssymb}

\makeatletter
\let\NAT@parse\undefined
\makeatother
\usepackage{hyperref}
\hypersetup{
    pdfborder={0 0 0},
    colorlinks=true,
    urlcolor=teal,
    linkcolor=teal,
    citecolor=teal,
    bookmarks=true,
    bookmarksnumbered=true
}

\graphicspath{{Figures/}}

\begin{document}
\title{Grasp Synthesis Matching From Rigid To Soft Robot Grippers Using Conditional Flow Matching}

\author { Tanisha Parulekar$^{1,2}$, Ge Shi $^{1*}$, Josh Pinskier$^{1}$, David Howard$^{1}$, Jen Jen Chung$^{2}$}

\maketitle

\begin{abstract}
A representation gap exists between grasp synthesis for rigid and soft grippers. Anygrasp~\cite{fang2023anygrasp} and many other grasp synthesis methods are designed for rigid parallel grippers, and adapting them to soft grippers often fails to capture their unique compliant behaviors, resulting in data-intensive and inaccurate models. To bridge this gap, this paper proposes a novel framework to map grasp poses from a rigid gripper model to a soft Fin-ray gripper. We utilize Conditional Flow Matching (CFM), a generative model, to learn this complex transformation. Our methodology includes a data collection pipeline to generate paired rigid-soft grasp poses. A U-Net autoencoder conditions the CFM model on the object's geometry from a depth image, allowing it to learn a continuous mapping from an initial Anygrasp pose to a stable Fin-ray gripper pose. We validate our approach on a 7-DOF robot, demonstrating that our CFM-generated poses achieve a higher overall success rate for seen and unseen objects ($34\%$ and $46\%$ respectively) compared to the baseline rigid poses ($6\%$ and $25\%$ respectively) when executed by the soft gripper. The model shows significant improvements, particularly for cylindrical ($50\%$ and $100\%$ success for seen and unseen objects) and spherical objects ($25\%$ and $31\%$ success for seen and unseen objects), and successfully generalizes to unseen objects. This work presents CFM as a data-efficient and effective method for transferring grasp strategies, offering a scalable methodology for other soft robotic systems.
\end{abstract}

% \footnotetext[1]{CSIRO Robotics, Australia}
% \footnotetext[2]{The University of Queensland, Australia}
% \thanks{Email: ge.shi@csiro.au}

% no keywords

\IEEEpeerreviewmaketitle

\section{Introduction}
Vision-guided object grasping for unstructured environments is a well-developed field, with state-of-the-art models enabling robust robotic manipulation~\cite{fang2023anygrasp, 11206484}. However, these models are predominantly designed for rigid parallel grippers. This focus creates a significant challenge for the field of soft robotics, where grippers leverage their inherent compliance to achieve exceptional flexibility and adaptability~\cite{rus2015design, pinskier2022bioinspiration}. Soft grippers excel at tasks that are difficult for their rigid counterparts, such as handling fragile objects~\cite{joseph2023jamming}, facilitating safe human-robot interactions~\cite{shi2024multi, pinskier2023automated, wang2025dexgrip}, and dynamically grasping poorly localized targets~\cite{shi2024design, greenland2025sograb}.

The primary obstacle to integrating these two fields is a fundamental mismatch in both grasp representation and control strategy. Rigid grasping methods rely on precise joint limit control and often fail to account for properties unique to soft grasping, such as time-based metrics and object deformation~\cite{fang2023anygrasp, newbury2023deep}. In contrast, soft grasping leverages the gripper's adaptability, often without requiring accurate finger modeling~\cite{choi2018learning}. This discrepancy means there is no consensus on how to generalize strategies from rigid to soft grippers, preventing the direct deployment of existing models.

This challenge is particularly evident for specific designs like the Fin-ray gripper. While its unique deformation abilities allow it to conform easily to various object shapes, its complex, non-linear behavior is difficult to model accurately and is simulation-intensive~\cite{pozzi2020hand, vatsal2022augmenting, shan2020modeling, de2021deep}. This modeling difficulty leads to a lack of accuracy when adapting grasp synthesis to real-world objects~\cite{de2021deep}, highlighting the need for a more efficient, less resource-intensive method to learn its grasping behaviors.

\begin{figure}
    \centering
    \includegraphics[width=1\linewidth]{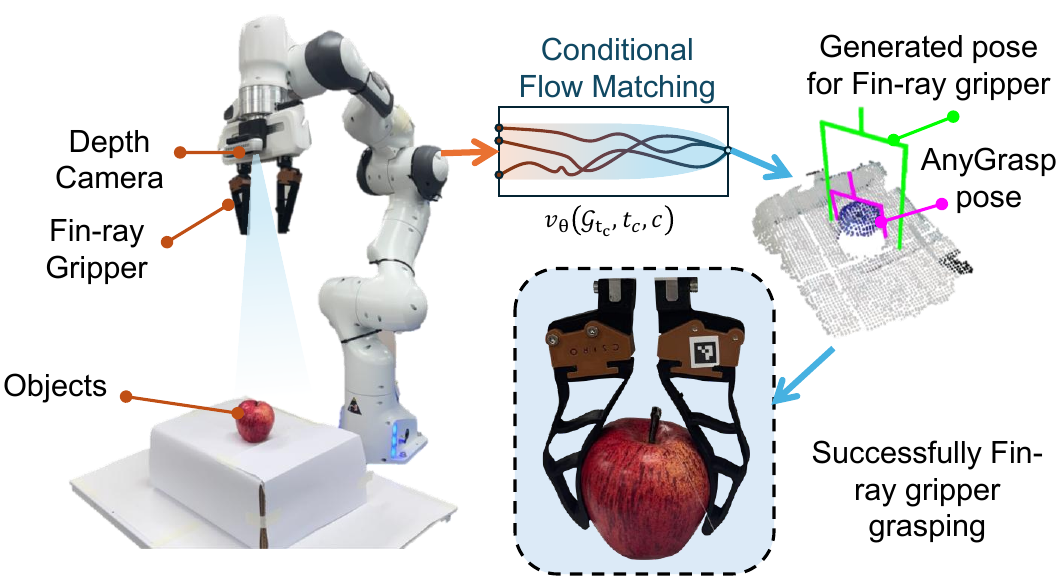}
    \caption{The CFM flow model will generate a successful grasp pose for a Finray gripper, given an AnyGrasp pose that is generated on an object.}
    \label{fig:IntroImage}
\end{figure}

To address this research gap, we deploy Conditional Flow Matching (CFM), a class of efficient and effective generative models for mapping data between different distributions~\cite{esser2024scalingrectifiedflowtransformers, tong2023conditional}. Compared to other generative models, CFM can achieve similar performance with less training data. Their application in robotics includes learning complex behaviors or waypoint planning~\cite{zhang2024affordance, chi2024diffusionpolicyvisuomotorpolicy}. Rigid grasping techniques rely on large dataset collection and capturing multiple views of grasping objects, which requires significant effort, training time and resource consumption. This paper, therefore, investigates the use of CFM as a less resource-intensive alternative. We propose a framework from data collection to mapping an Anygrasp rigid-gripper grasping pose to a successful grasp pose for a soft Fin-ray gripper, as illustrated in Fig.~\ref{fig:IntroImage}.

The main contribution of this work is the development of the CFM model to transfer grasp poses, originally synthesized for rigid grippers, to a soft Fin-ray gripper. To accomplish this, we establish a data collection pipeline to generate the required paired grasp data and subsequently implement and analyze the effectiveness of the CFM as a grasp synthesis technique for the soft gripper.

\section{Related Work} \label{related}
% Briefly introduce in a few sentences Fin-ray grippers and their relevance. 
\subsection{Grasp Synthesis} \label{grasp_synth}
%AnyGrasp, synthesis datasets that generalise across different grippers, where grasp synthesis is for soft grippers. 
Grasp synthesis is the process of generating grasp hypotheses and contact models for either fixture (maintaining equilibrium against disturbances) or dexterous manipulation (moving an object in a specific way)~\cite{newbury2023deep}. A state-of-the-art model for rigid grippers is Anygrasp, which utilizes scene reconstruction and additional geometric labels to improve grasp stability and association over time~\cite{fang2023anygrasp}. It remains unconfirmed whether the assumptions underlying rigid parallel grippers are generalizable to soft robot grippers, as passively actuated systems may not exhibit the same behaviors or benefit from the same ideal orientations. For example, Anygrasp and other methods for parallel grippers assume antipodal grasping points are ideal and abstract key parameters such as the width and depth of the gripper.

While other efforts have been made to transfer grasps across grippers with varying numbers of fingers, different widths and finger depths, or even different actuation methods (e.g., parallel jaw vs. vacuum)~\cite{sarmientogripper, gilles2023metagraspnetv2}, these approaches rely on individual joint control. They introduce fixed relationships between parameters like finger depth and grasp width---a paradigm that is incompatible with most soft grippers.
%\subsection{Soft Grasping} \label{grasp_soft}
Soft grasping methods still heavily rely on rigid grasp synthesis. Early research demonstrated that sufficient accuracy could be achieved by adapting learning-based methods from rigid grippers, using a simplified open-close control sequence to leverage the gripper's compliance~\cite{choi2018learning}. However, such simplified control schemes can disregard the unique behaviors of soft grippers, which are critical for selecting a successful grasp. This reliance on rigid-body techniques persists in more recent work. For instance, some approaches account for the hand closure of a soft hand but still use rigid grasp generators like DexNet2.0~\cite{pozzi2020hand}, while others use datasets designed for rigid grippers, like the Cornell dataset, to generate commands for a soft hand~\cite{vatsal2022augmenting}. This continued reliance on rigid grasp synthesis has created a representation gap, failing to capture the unique behaviors and capabilities of soft grippers.

\subsection{Flow Models} \label{flow models}
Generative models offer a novel perspective to bridge the gap between rigid and soft gripper grasp synthesis. We leverage generative models, which learn a probabilistic mapping between two data distributions. One prominent approach, Normalizing Flows (NFs) and their conditional variants (CNFs), can learn complex, invertible transformations by parameterizing them as continuous-time Ordinary Differential Equations (ODEs)~\cite{tong2023improving}. Despite their flexibility in density estimation, CNFs have historically been hindered by difficulties in training and scaling~\cite{onken2021ot, tong2023improving}. An alternative, diffusion models, represents the current state-of-the-art in many generative tasks but often requires numerous network passes to generate a high-quality sample, leading to long inference times~\cite{lipman2022flow, liu2022rectified}.

Addressing these limitations, Conditional Flow Matching (CFM) has recently emerged as a highly efficient training method for such generative models~\cite{tong2023conditional}. By directly learning the transformation between two domains---in our case, from rigid gripper grasps to soft gripper grasps---CFM provides an accurate mapping with significant computational efficiency. Flow-based methods have recently been applied to various robotics challenges, including motion planning, affordance-based control, and multi-support manipulation~\cite{rouxel2024flow, zhang2024affordance, nguyen2025flowmp, ye2024efficient}. This combination of accuracy and efficiency makes CFM particularly suitable for learning the complex conditional transformations required in soft robotics.

\section{Methodology} \label{design}
\begin{figure*}[h!]
    \centering
    \includegraphics[width=\linewidth]{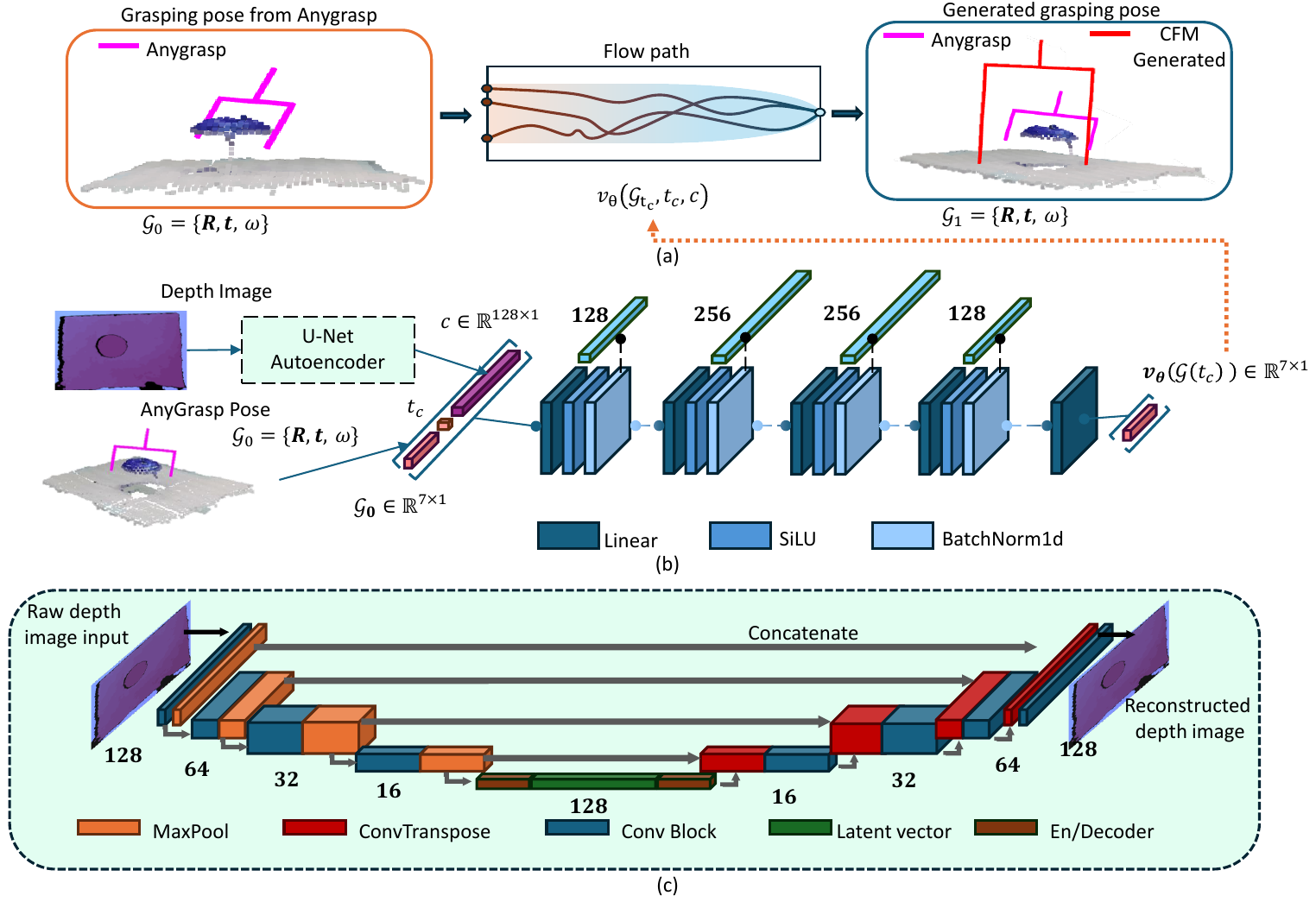}
    \caption{An overview of the Conditional Flow Matching (CFM) framework for grasp synthesis mapping. (a) The core concept of the CFM model that learns a continuous transformation (a ``flow path") from an initial rigid gripper pose generated by Anygrasp ($\mathcal{G}_{Anygrasp}$) to a target soft gripper pose ($\mathcal{G}_{\text{CFM}}$). This transformation is guided by a learned, conditional velocity field $\mathbf{v}_\theta$. (b) This shows the architecture of the feed-forward MLP that parameterizes the velocity field $\mathbf{v}_\theta$. It takes the current grasp pose $\mathcal{G}(t_c)$, the progression parameter $t_c$, and the scene condition vector $\mathbf{c}$ as input to predict the direction of the flow. (c) This depicts the U-Net Autoencoder used to generate the condition vector $\mathbf{c}$ at the bottle-neck. It processes a raw depth image, compressing it into a latent vector that captures the object's geometry.}
    \label{fig:flowoverview}
\end{figure*}
We represent the state of a grasp pose as $\mathcal{G} = \{ \mathbf{R}, \mathbf{t}, w \}$, where \(\mathbf{R} \in \mathbb{R}^{4 \times 1}\) represents the gripper orientation, and \(\mathbf{t} \in \mathbb{R}^{3 \times 1}\) represents the position of the center of the gripper, and $w$ is the gripper opening width. This representation covers all the degrees of freedom for a parallel gripper and is also referred to as the 7-DOF grasp configuration. The gripper opening width is assumed to be a constant of the max opening width, therefore it is not included in the grasp configration. In this paper, we use the notation $\mathcal{P}$ to represent the partial-view point cloud from a depth camera associated with a depth image $\mathcal{D}$, and $\mathcal{E}$ to represent the environment including the robot and objects. Hence a robot grasping scene is defined as $\mathcal{S} = \{\mathcal{E}, \mathcal{P}, \mathcal{D}, \mathcal{G}\}$.

The approach, leveraging CFM, is introduced to achieve the grasp synthesis mapping. As shown in Fig.~\ref{fig:flowoverview}(a), the core of the CFM is to model the transformation from an Anygrasp rigid gripper grasping pose $\mathcal{G}_{Anygrasp}$ to the soft gripper grasping pose $\mathcal{G}_{\text{CFM}}$ as the solution to a conditional Ordinary Differential Equation (ODE). The transformation is governed by a parameterized vector field conditioned on the condition input $c$:
\begin{equation}
    \frac{d\mathcal{G}(t_{c})}{dt_{c}} = \mathbf{v}_\theta(\mathcal{G}(t_{c}),t_{c},\mathbf{c}), \; \mathcal{G}(0) = \mathcal{G}_{Anygrasp}, \; \mathcal{G}(1) = \mathcal{G}_{\text{CFM}},
    \label{traject_ODE}
\end{equation}
where $t_c \in [0,1]$ is a progression parameter in the ODE \(\frac{d\mathcal{G}(t_c)}{dt_c} = \mathbf{v}_\theta(\mathcal{G}(t_c),t_c,\mathbf{c})\), and \(\mathbf{v}_\theta\) is a time-dependent vector field parameterized by an MLP with trainable parameters ($\theta$). The trajectory from \(\mathcal{G}_{Anygrasp}\) to \(\mathcal{G}_{\text{CFM}}\) is obtained by integrating \eqref{traject_ODE} over $t_c$ from 0 to 1~\cite{rezende2015variational}.
\begin{figure*}[h!]
    \centering
    \includegraphics[width=0.9\linewidth]{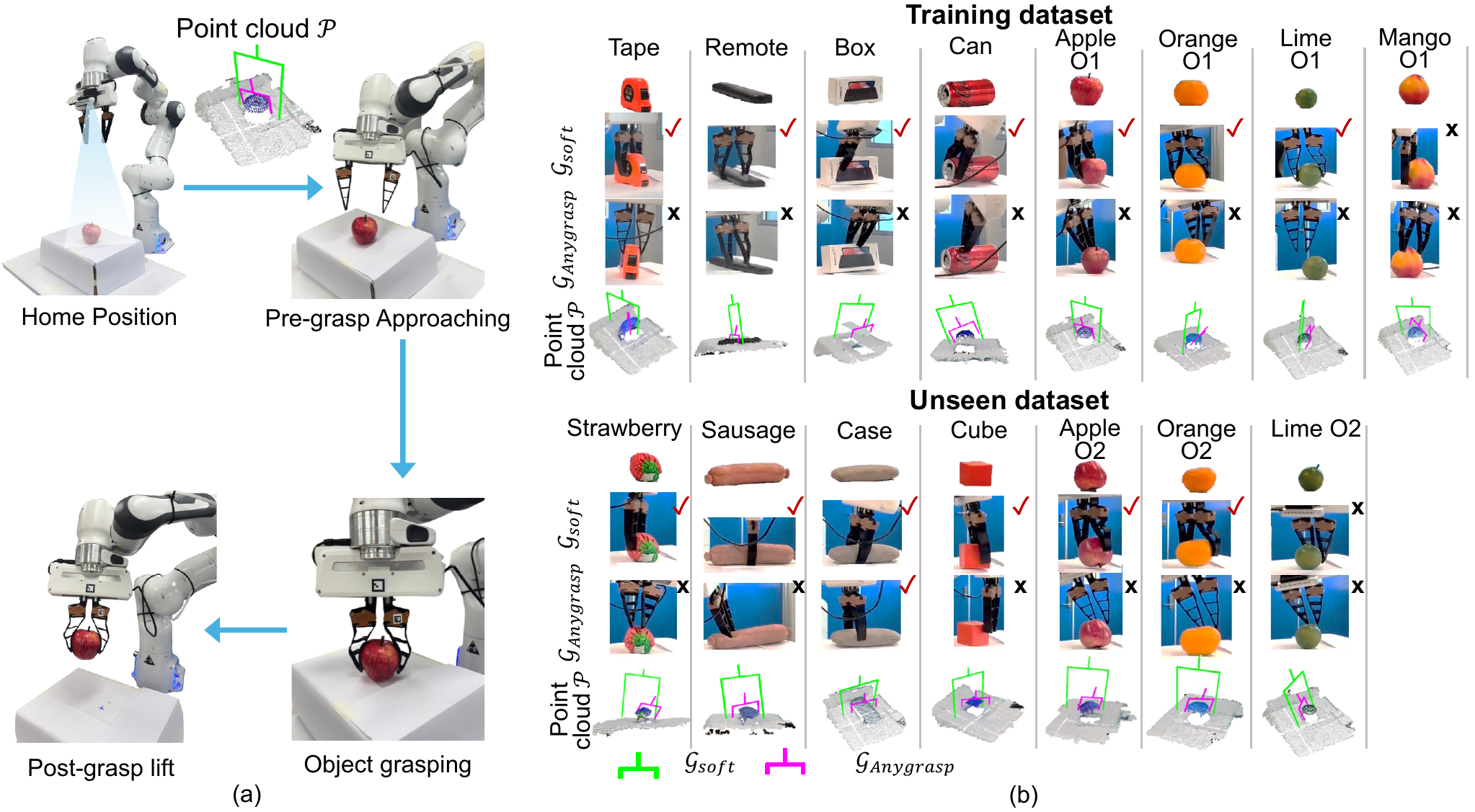}
    \caption{The experimental setup and dataset used for learning the grasp pose transformation. (a) This shows the data validation pipeline for validating the CFM model, which involves moving the robot from a home position to a pre-grasp pose derived from \(\mathcal{G}_{Anygrasp}\) or \(\mathcal{G}_{\text{CFM}}\), executing the grasp on the object, and lifting it. (b) This section displays examples from the training and unseen datasets for a variety of objects. Each entry includes the grasp execution, the corresponding point cloud ($\mathcal{P}$), the initial rigid gripper pose from AnyGrasp (\(\mathcal{G}_{Anygrasp}\), shown in magenta), and the manually adjusted, successful soft gripper pose (\(\mathcal{G}_{\text{CFM}}\), shown in green).}
    \label{fig:exp_setup}
\end{figure*}
The vector field $\mathbf{v}_\theta$ is learned by a fully-connected, feed-forward MLP, whose architecture is detailed in Fig.~\ref{fig:flowoverview}(b). The input to this network consists of the interpolated grasp pose $\mathcal{G}(t_c)$, the progression parameter $t_c$, and the condition vector $\mathbf{c}$. For network processing, the grasp pose $\mathcal{G}$ is represented as a 7-dimensional vector, where the orientation $\mathbf{R}$ is converted to a 4D quaternion and concatenated with the 3D position vector $\mathbf{t}$. The MLP architecture consists of four hidden layers (sizes 128, 256, 256, 128) with SiLU activations, batch normalization, and a final ReLU activation.

Under a grasping scene $\mathcal{S} = \{\mathcal{E}, \mathcal{P}, \mathcal{D}, \mathcal{G}\}$, the grasp pose $\mathcal{G}_{Anygrasp}$ is generated according to the partial-view point cloud $\mathcal{P}$ extracted from depth image $\mathcal{D}$. To allow the mapping to adapt to different objects and scenes, the condition vector $\textbf{c}$ encodes the scene's geometry. As shown in Fig.~\ref{fig:flowoverview}(c), the U-Net autoencoder \cite{DBLP:journals/corr/RonnebergerFB15} was used to generate an encoding for $\textbf{c}$. The network takes the raw depth image from the current scene as input and processes it through a U-net structure to compress the visual information into a 128-dimensional latent vector at the bottle-neck. This latent vector, which captures the essential geometry of the object, then serves as the condition $\mathbf{c} \in \mathbb{R}^{128 \times 1}$ for the CFM model. The autoencoder was pre-trained on a dataset of 865 depth images of various objects to effectively learn this compressed representation.

Fig.~\ref{fig:flowoverview}(b) shows the structure of the CFM. Given a dataset pair of the grasp scenes for Anygrasp $\mathcal{S}_{Anygrasp} = \{\mathcal{E}, \mathcal{P}, \mathcal{D}, \mathcal{G}_{Anygrasp}\}$ and a demonstrated successful Fin-ray gripper grasp $\mathcal{S}_{Soft} = \{\mathcal{E}, \mathcal{P}, \mathcal{D}, \mathcal{G}_{\text{CFM}}\}$, the goal of CFM is to learn a velocity field that aligns with the transition from the Anygrasp grasp pose $\mathcal{G}_{Anygrasp}$ to a target soft grasp pose $\mathcal{G}_{\text{CFM}}$. The pair of grasp scenes will be generated for an object $\mathcal{E}$,  in the context of a point cloud  $\mathcal{P}$ . A depth image $\mathcal{D}$ with conditions as $\textbf{c}_{Anygrasp} = \textbf{c}_{soft}$ will be associated with the scene context, and will link the Anygrasp pose and demonstrated Fin-ray gripper grasp pose.  Both grasp poses will be for the same robot arm with the Fin-ray gripper attached. Given the data $(\mathcal{G}_{Anygrasp},\mathcal{G}_{\text{CFM}})$ under the same conditions $\textbf{c}$, an interpolated state $\mathcal{G}_{t_c}$ at a random progression time $t_c\sim U(0,1)$ is defined as:
\begin{equation}
    \mathcal{G}_{t_c} = (1-t_c)\mathcal{G}_{Anygrasp} + t_c\mathcal{G}_{\text{CFM}}.
    \label{interpolate-path}
\end{equation}
The target velocity field, representing the ideal direction of transformation, is:
\begin{equation}
    \mathbf{u}_{t_c} = \mathcal{G}_{\text{CFM}} - \mathcal{G}_{Anygrasp}.
    \label{direct-transform}
\end{equation}
Meanwhile the training loss is formulated as the mean squared error between the predicted and target velocity fields:
\begin{equation}
    \mathcal{L}(\theta) = \mathbb{E}_{t_c,\mathcal{G}_{Anygrasp},\mathcal{G}_{\text{CFM}},\mathbf{c}}[\|\mathbf{v}_\theta(\mathcal{G}(t_c),t_c,\mathbf{c})-\mathbf{u}_{t_c}\|^2].
    \label{loss-train}
\end{equation}
By defining the loss function, the model tends to learn a smooth, conditional mapping from $\mathcal{G}_{Anygrasp}$  to $\mathcal{G}_{\text{CFM}}$. After training, given a scene $\mathcal{S}_{Anygrasp} = \{\mathcal{G}_{Anygrasp}, \textbf{c}\}$, the target grasping synthesis $\mathcal{G}_{\text{CFM}}$ is predicted by solving:
\begin{equation}
   \mathcal{G}_{\text{CFM}} =\mathcal{G}_{Anygrasp}\,+\,\int_0^{1}\mathbf{v}_\theta(\mathcal{G}(t_c),t_c,\mathbf{c}).
    \label{sim2real-velocity_field}
\end{equation}
This integral is numerically approximated using the ODE solver (e.g., Dormand-Prince method)~\cite{tong2023conditional}. The CFM framework learns the velocity field $v_\theta$ using a fully-connected, feed-forward MLP. The input to the network is the current state vector $\mathcal{G}(t_c)\in\mathbb{R}^N $ of the observation data, the flow-time parameter $t_c$, and the condition vector $c\in\mathbb{R}^{128 \times 1}$. Our system was run on a laptop with 11th Gen Intel i9-11950H CPU and an RTX 2000 Ada GPU.
\section{Experiments} \label{transfer_eval}
\subsection{Paired Data Collection}
A custom experimental platform was built for paired grasping dataset collection and validation, as shown in Fig.~\ref{fig:exp_setup}(a). The setup consisted of a 7-DOF Franka Emika Panda manipulator with an end-effector-mounted Intel RealSense D415 depth camera and a Fin-ray gripper for the grasping task. The data was generated using a pick-and-place pipeline. For each trial, the depth camera captured a partial-view depth image ($\mathcal{D}$) from a top-down perspective. Anygrasp was run to generate the associated scene point cloud  ($\mathcal{P}$) and synthesise a sequence of 20 ranked grasping poses ($\mathcal{G}_{Anygrasp}$) suitable for a rigid gripper. During data collection, the robot first moved to a pre-grasp pose, at a $15\,\text{cm}$ offset along the approach vector of a selected Anygrasp pose. This procedure ensured the end-effector maintained the same orientation and approach direction as the original predicted grasp. From this pre-grasp pose, the grasp was manually adjusted to find a successful pose, $\mathcal{G}_{\text{CFM}}$, for the Fin-ray gripper.

Each collected data pair consists of the initial Anygrasp pose $\mathcal{G}_{Anygrasp}$ and the corresponding successful, manually-adjusted Fin-ray gripper pose $\mathcal{G}_{\text{CFM}}$, conditioned on the same depth image $\mathcal{D}$. The $\mathcal{G}_{Anygrasp}$ poses form the initial data distribution, while the successful $\mathcal{G}_{\text{CFM}}$ poses form the target distribution for our CFM model. To ensure a robust dataset, both higher- and lower-ranking Anygrasp poses were selected for inclusion. By collecting these pairs, the CFM model learns to map a given Anygrasp pose to an effective Fin-ray grasp. For each training object shown in Fig.~\ref{fig:exp_setup}(b), we collected 15 data pairs, with the adjusted grasps demonstrating top-down approaches at varying depths ($1\,cm$ to $7\,cm$, in $1\,cm$ increments).
\subsection{Validation}
To validate the CFM model, the pick-and-place pipeline was employed across a total of 15 scenes involving 12 distinct objects, which included the training dataset and unseen objects. The unseen objects also included an apple, orange, and lime placed in upside-down orientations. During validation, the Panda manipulator equipped with a pair of Fin-ray grippers executed the original $\mathcal{G}_{Anygrasp}$ poses and the CFM-generated $\mathcal{G}_{\text{CFM}}$ poses. To better demonstrate the improvement offered by the CFM model, middle-ranking $\mathcal{G}_{Anygrasp}$ poses were intentionally selected for the baseline comparison. In each experimental trial, the manipulator approached the pre-grasp pose for a given $\mathcal{G}_{Anygrasp}$ or $\mathcal{G}_{\text{CFM}}$, closed the gripper, and lifted the object. A grasp was considered successful if the object could be held stable for 5 seconds after being lifted and placed back on the platform.
        
\section{Results} \label{results}
\subsection{Training and Validation Results}
The CFM model was trained for 2000 epochs on 254 paired grasp poses from 8 different objects, using an 80/20 training and validation split. The associated U-Net autoencoder was trained for 5000 epochs on a dataset of 865 depth images. The total training time was approximately 1 minute for the CFM model and 35 minutes for the autoencoder. The trained model demonstrated a strong ability to replicate the manually adjusted poses, with minimal differences observed between the demonstrated and the predicted Fin-ray gripper poses.

\begin{figure}[t]
    \centering
    \includegraphics[width=0.7\linewidth]{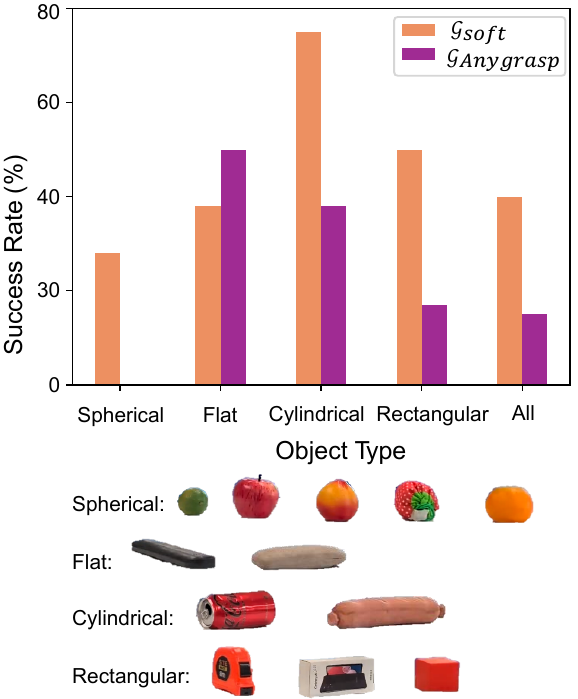}
    \caption{Benchmarking results comparing the grasp success rates of the CFM-generated soft gripper poses ($\mathcal{G}_{{soft}}$) against the original AnyGrasp poses ($\mathcal{G}_{{Anygrasp}}$).}
    \label{fig:ValidationResults}
\end{figure}

To quantitatively evaluate its effectiveness, the CFM model's grasp synthesis ($\mathcal{G}_{\text{CFM}}$) was benchmarked against the baseline Anygrasp synthesis ($\mathcal{G}_{Anygrasp}$) on both seen and unseen objects. The results, summarized in Fig.~\ref{fig:ValidationResults}, shows that  the CFM model achieved a higher overall success rate of approximately 38\% compared to Anygrasp's 15\%. For seen and unseen objects, the CFM model achieved higher success than Anygrasp for seen (34\% and 6\% respectively) and unseen objects (46\% and 25\% respectively). Performance also varied by object geometry. For cylindrical objects, the CFM model showed the most significant improvement, achieving a success rate of 50\% and 100\% respectively for the seen and unseen objects, which were better than Anygrasp (0\% and 75\% respectively). The CFM model also outperformed the baseline on rectangular objects, achieving 50\% for both seen and unseen objects compared to Anygrasp (25\% and 0\% respectively). For spherical objects, the model yielded success rates of approximately 25\% and 31\% for seen and unseen objects respectively, whereas the baseline Anygrasp method failed completely.  Flat objects were the only category in unseen objects where the baseline Anygrasp model performed significantly better, achieving a success rate of 100\% for both seen and unseen objects, while the CFM model's success rate was 25\% and 50\% respectively. \par 
% the CFM model achieved a higher overall success rate of approximately 38\% compared to Anygrasp's 15\%. The model's performance varied by object geometry. For cylindrical objects, the CFM model showed the most significant improvement, achieving a success rate of over 75\%, almost double that of the baseline's ~38\% success rate. For spherical objects, the model yielded a success rate of approximately 34\%, whereas the baseline Anygrasp method failed completely. The CFM model also outperformed the baseline on rectangular objects, with a success rate of nearly 50\% compared to Anygrasp's ~28\%. Flat Objects were the only category where the baseline Anygrasp model performed better, achieving a success rate of nearly 50\%, while the CFM model's success rate was just under 40\%. 
\par
% The CFM model trained on the apple and orange was tested on the lime to understand the performance for unseen objects. The model had a 20\% success rate with the unseen object. 
% \begin{figure}[H]
%     \centering
%     % \includegraphics[width=1\linewidth]{}
%     \caption{Testing results for CFM model.}
%     \label{fig:ValidationResults}
% \end{figure}\par

\section{Discussion} \label{section: discussion}
Overall, the CFM model successfully synthesized grasps across different objects for the Fin-ray gripper. The experimental results suggest that the CFM model is capable of learning a mapping that compensates for the morphological and compliance differences between a rigid parallel gripper and a soft Fin-ray gripper. The model's primary adaptations---increasing grasp depth, correcting tilted orientations, and re-centering grasps on planar surfaces---are consistent with the known operational principles of soft grippers, which favor distributed forces and envelopment over precise, high-pressure contacts.

During the data collection stage, the primary adaptation was the adjustment of grasp depth as well as suboptimal orientations. Grasps synthesized by Anygrasp are often shallow and the contact points are in the middle of the object to ensure the stable grasp for a rigid pinch gripper. This often requires the gripper to be highly precise, with high-pressure point contacts. In contrast, a deeper grasp is needed for an object to allow the adaptation of the Fin-ray gripper actuation during grasping. This strategic shift allows the longer fingers of the Fin-ray gripper to maximize the surface contact and leverage their enveloping capability, which is crucial for stability with soft grippers.

Furthermore, the model learned to enhance stability by correcting suboptimal orientations proposed by the baseline. Anygrasp often produced tilted grasps or targeted the corners of cubic objects---strategies viable for rigid grippers that can cage an object. For a soft gripper, such approaches can lead to uneven force application and result in slips. The CFM model consistently re-adjusts these poses to more stable, top-down, and centered orientations, suggesting it developed a preference for balanced force distribution, a key principle in successful grasping.

It is worth noting that the baseline Anygrasp model performed better on flat objects. This limitation may be caused by bias during the data collection. The manually adjusted grasps predominantly featured top-down approaches, which are effective for many shapes but less optimal for flat objects where a pinching grasp from the side can be more stable. The learned top-down strategy was more effective than Anygrasp for the remote as it increased the grasp depth. However, for the glasses case a pinching grasp from the side may have been a more effective strategy. This indicates that while the model learns from the provided data efficiently, its performance is necessarily constrained by the scope of the demonstrated strategies.

The success of the model can also be partially attributed to the inclusion of high- and low-ranking poses from Anygrasp. The diversity allowed the model to learn a robust corrective mapping rather than a simple one-to-one transfer. This likely contributed to its ability to transform poses considered suboptimal for a rigid gripper into successful grasps for the soft gripper. This suggests that the model is not merely memorizing good grasps but is learning a conditional policy for what constitutes a stable grasp given the Fin-ray gripper's physical properties.

The model's ability to successfully generalize to unseen objects underscores a key advantage of the CFM framework: its generative feature. Unlike a simple regression model that might overfit to the specific geometries in the training set, the CFM learns a continuous, conditional vector field that represents the principles of transformation. The U-Net autoencoder is critical for this process. By mapping the geometries of unseen objects to a familiar region within its latent space, it provides a consistent conditioning vector $\mathbf{c}$ that allows the learned flow to be applied effectively, even to objects the model has never seen before. Therefore, the success on unseen objects indicates that the model has not merely memorized the training pairs but has captured the underlying conditional policy for transforming a rigid grasp into a viable soft grasp. This demonstrates the potential of flow-based models to create robust, generalizable solutions for robotic manipulation tasks that can adapt to novelty in the environment.

\section{Conclusion} \label{section: conclusion}
In summary, this work demonstrates that CFM is a beneficial and data-efficient method for adapting Anygrasp-synthesized rigid grasps for use with soft Fin-ray grippers, presenting a mapping methodology that could be extended to other soft grippers. Benchmarking experiments, however, revealed areas for improvement, as common failure modes included grasps that were off-center, positioned on the side of the object, too shallow, or excessively tilted, sometimes resulted in slips. This suggests that future work could focus on integrating grasp centering methods or improving object geometry perception to enhance the model's robustness.
\bibliographystyle{IEEEtran}

\bibliography{bib_tanisha}

\end{document}